\documentclass[runningheads]{llncs}
\usepackage[T1]{fontenc}
\usepackage{graphicx}
\usepackage{booktabs}
\usepackage[misc]{ifsym}

\usepackage{amssymb}
\usepackage{amsmath}
\usepackage{multirow}   
\usepackage[table]{xcolor} 
\definecolor{bestc}{HTML}{D1E8E2}  
\definecolor{secondc}{HTML}{FFF2CC}
\usepackage[ruled,vlined]{algorithm2e}
\definecolor{mycite}{RGB}{0,0,204}
\definecolor{myref}{RGB}{192,0,0}
\usepackage{hyperref}
\hypersetup{
    colorlinks=true,
    linkcolor= myref,
    citecolor= mycite
}

\begin{document}

\title{Imputation Meets Clustering: Exploiting Latent Subgroup Structure for Missing Data Recovery}

\titlerunning{Cluster-Aware Generative Imputation}

\toctitle{Imputation Meets Clustering: Exploiting Latent Subgroup Structure for Missing Data Recovery}
\tocauthor{Chuyao Zhang, E Li, Taochen Chen, Yiqun Zhang, Yuzhu Ji, Shuping Zhao, Peng Liu, Yiu-ming Cheung}

\author{Chuyao Zhang\inst{1} \and
E Li\inst{1} \and
Taochen Chen\inst{1} \and
Yiqun Zhang\inst{1} (\Letter) \and \\
Yuzhu Ji\inst{1} \and 
Shuping Zhao\inst{1} \and
Peng Liu\inst{1} \and
Yiu-ming Cheung\inst{2}}

\authorrunning{C. Zhang et al.}

\institute{Guangdong University of Technology, Guangzhou, China \\
\email{\{zhangchuyao, lie1, chentaochen1\}@mails.gdut.edu.cn}\\
\email{\{yqzhang, yuzhu.ji, shupingzhao, liupeng\}@gdut.edu.cn}
\and
Hong Kong Baptist University, Hong Kong, China \\
\email{ymc@comp.hkbu.edu.hk}
}

\maketitle 

\begin{abstract}
    Missing data is prevalent in practical applications, making effective imputation an essential preprocessing step for downstream analysis. Real-world datasets often exhibit complex latent structures composed of multiple subgroups with distinct distributions. However, existing methods often overlook such population heterogeneity. Without explicit structural guidance, these methods tend to produce generic estimates that blur subgroup boundaries and lack instance-level fidelity. While incorporating subgroup information offers a remedy, it faces a circular dependency: reliable subgroup identification requires complete data, while data completion is the imputation objective itself. To resolve this, we propose CAGI (Cluster-Aware Generative Imputation), a framework that reformulates clustering and imputation as a mutually reinforcing co-optimization process. CAGI employs a ``Partition-Guide-Restore'' strategy where dynamic cluster assignments act as local priors to condition a Generative Adversarial Network. An iterative feedback loop is established to progressively refine both cluster structures and imputed values toward faithful subgroup distributions. To ensure distributional stability, CAGI further employs a multi-level optimization objective combining instance-level reconstruction with distribution-level regularization. Extensive experiments on 14 benchmark datasets with 15 representative baselines demonstrate the superiority of CAGI.
    The source code is available at: \textcolor{magenta}{\url{https://github.com/supercocachii/CAGI}}.

\end{abstract}

\section{Introduction} \label{sec:intro}
    \begin{figure}[t]
        \centering
        \includegraphics[width=1\textwidth]{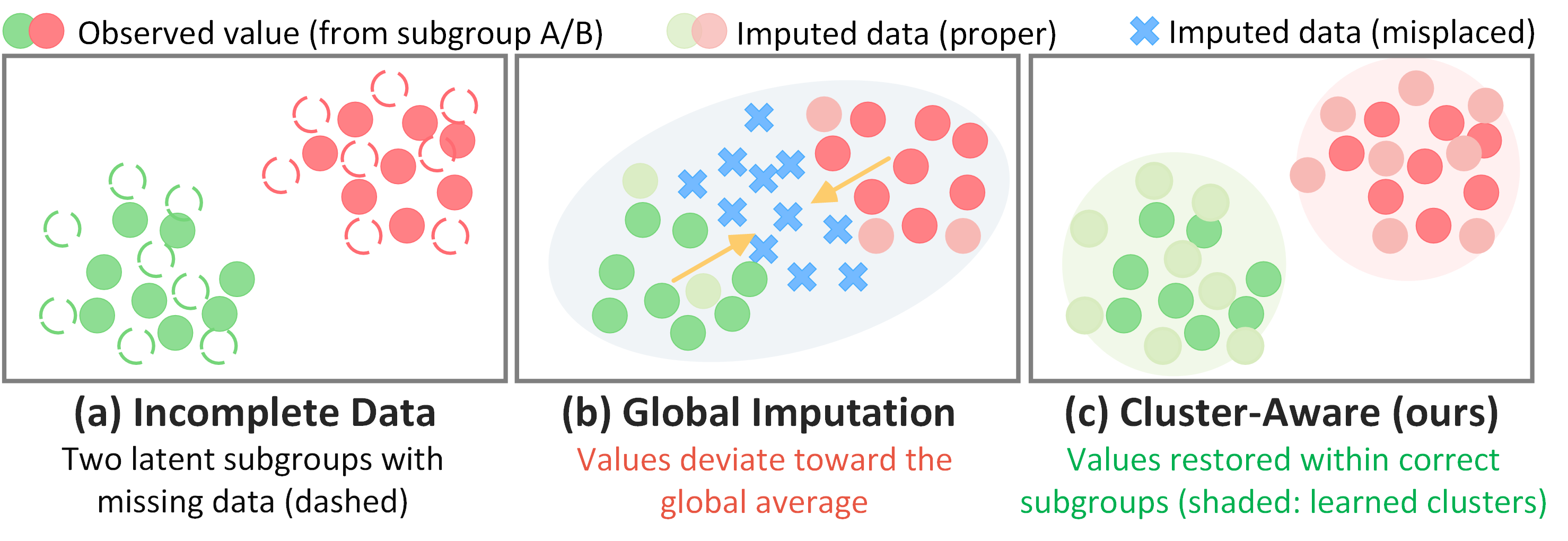} 
        \caption{
        Motivating example of imputation under subgroup heterogeneity in an unsupervised setting. Subgroup colors are shown only for illustration. (a) Incomplete data with two latent subgroups whose memberships are unknown. (b) Global imputation, agnostic to latent structure, deviates toward the global average and produces misplaced values. (c) CAGI discovers subgroup structure (shaded regions indicate learned cluster boundaries) and restores correct regions.}
        \label{fig:motivation}
    \end{figure}
    
    Missing values are pervasive in real-world datasets, arising from sensor failures~\cite{zhang2023time}, human entry errors, privacy protection mechanisms, and other factors~\cite{survey}. Such incompleteness poses fundamental risks to downstream learning tasks, degrading classification accuracy, producing unreliable clustering, and introducing bias into automated decision systems~\cite{impact}.
    To mitigate these risks, \textit{imputation}, the process of estimating plausible values for unobserved entries based on available information, has been a widely used preprocessing step. 
    Crucially, imputation is not merely about filling in blanks; its fundamental goal is to recover the true latent distribution from partially observed samples, thereby providing a reliable foundation for subsequent analysis.

    Missing data mechanisms are typically categorized into three types~\cite{mean}: (1) Missing Completely At Random (MCAR), where missingness is independent of all data; (2) Missing At Random (MAR), where missingness depends only on observed values; and (3) Missing Not At Random (MNAR), where missingness depends on the unobserved values themselves.
    Imputation methods have advanced substantially over the past decade. Following common practice in recent imputation benchmarks~\cite{grape,remasker}, this paper focuses on the MCAR setting.    
    
    Early statistical approaches such as Mean/Mode substitution~\cite{mean} and EM algorithm~\cite{em} offer simple and interpretable solutions, but rely on parametric or linear assumptions that limit their expressiveness.
    Iterative conditional approaches such as MICE~\cite{mice} and MissForest~\cite{missforest} estimate missing values through chained conditional models or Random Forests, but may face scalability challenges on high-dimensional data with complex inter-variable structures.
    More recently, deep generative models offer more expressive alternatives. GAN-based methods such as GAIN~\cite{gain} model the data distribution through adversarial training, VAE-based methods such as MIWAE~\cite{miwae} learn latent representations via variational inference, and diffusion-based methods such as TabCSDI~\cite{tabcsdi} and DiffPuter~\cite{diffputer} reconstruct data through iterative denoising. These methods demonstrate a superior ability to capture complex feature correlations.

    Despite their power, existing methods often treat the dataset as a monolithic entity, implicitly learning a smooth, unimodal manifold over all samples. This assumption overlooks the latent population heterogeneity prevalent in practice. 
    Real-world datasets are typically composed of multiple subgroups with distinct statistical characteristics. For instance, patients in clinical diagnostics may belong to distinct disease subtypes, while customers in marketing databases fall into diverse demographic segments.
    When a unified model is trained on such distributionally heterogeneous data without explicit structural guidance, the learned distribution conflates multiple modes into a global average. 
    As illustrated in Figure~\ref{fig:motivation}, consider a dataset containing two subgroups with missing values (Figure~\ref{fig:motivation}(a)). A global imputation model, agnostic to this latent structure, generates estimates reflecting the marginal distribution averaged across subgroups (Figure~\ref{fig:motivation}(b)). 
    These imputed values reside in regions of the feature space that are ``plausible on average'' but ``incorrect for any specific subgroup''.
    This leads to two key issues: \textit{distributional shift} occurs when imputed values deviate from each subgroup's true conditional distribution, and \textit{boundary violation} occurs when imputed values cross subgroup boundaries and distort the subgroup structure. Such errors are not merely theoretical concerns. For instance, in clinical applications, imputing biomarker values without accounting for disease subtypes can obscure important diagnostic signals.
    
    A natural remedy is to first identify subgroups via clustering and then impute within each subgroup. However, this straightforward strategy encounters a fundamental circular dependency: reliable subgroup identification requires complete data, while complete data is the very objective of imputation.
    Prior two-stage approaches~\cite{first,ccgain} adopt a cluster-then-impute pipeline that treats cluster assignments as one-off preprocessing outputs. Since errors in the initial cluster assignments are irreversibly fixed, such methods cannot correct the subgroup structure during the learning process.

    This paper, therefore, proposes CAGI (Cluster-Aware Generative Imputation) to reformulate clustering and imputation as a coupled co-optimization process. The key insight lies in the fact that identifying latent structures and recovering missing entries are coupled objectives that enhance each other through iterative feedback.
    Following a ``Partition-Guide-Restore'' strategy, CAGI first bootstraps subgroup discovery via missing-tolerant clustering that provides initial cluster assignments even when data is substantially incomplete.
    Then an adversarial generator incorporates these assignments as conditioning signals, narrowing the generative search space to synthesize missing values that adhere to the statistical patterns of specific subgroups.    
    Finally, it feeds refined imputations back to periodic reclustering, establishing an iterative imputation-reclustering-reimputation cycle that progressively improves both clustering and imputation. Accordingly, CAGI produces plausible imputed values that are in line with the specific subgroup structure (Figure~\ref{fig:motivation}(c)).   
    To guard against distributional fragmentation caused by inaccurate early clusters, CAGI further employs a multi-level optimization objective: instance-level constraints (adversarial training combined with reconstruction loss) ensure element-wise accuracy, while a distribution-level optimal transport regularization based on Sinkhorn divergence enforces global geometric consistency. The main contributions are:
    \begin{itemize}
        \item \textbf{Subgroup-Conditional Generative Strategy.} 
        A cluster-conditioned generation approach that decomposes the complex global distribution into compact subgroup-specific manifolds, reducing generative uncertainty in high-dimensional regimes.        
        \item \textbf{Alternating Optimization with Self-Correction.}
        We develop an iterative imputation-clustering loop that treats cluster assignments as tunable latent variables, anchored by reconstruction constraints on observed entries to progressively rectify early-stage clustering deviations.         
        \item \textbf{Multi-Level Objective for High-Fidelity Recovery.} 
        A composite objective combining instance-level reconstruction with distribution-level regularization is designed, preserving distinct local patterns while simultaneously maintaining a smooth global data topology.
    \end{itemize}

\section{Related Work} \label{sec:rw}
    Deep learning-based imputation and unsupervised data analysis techniques that are highly relevant to CAGI and the targeted mixed data are discussed below.

    \subsection{Deep Learning Approaches for Imputation}
    Deep generative models address the limitations above by learning to approximate the underlying data distribution.
      
    \textbf{GAN-based methods.} GAIN~\cite{gain} adapts the GAN~\cite{gan} framework to imputation by training a generator to fill missing entries and an element-wise discriminator to identify which components were imputed. MisGAN~\cite{misgan} extends this by jointly learning a data generator and a mask generator to model complex missingness patterns. While computationally efficient, GAN-based methods are susceptible to training instability~\cite{gan-1,gan-2}.

    \textbf{VAE-based methods.} Variational Autoencoder approaches optimize the evidence lower bound for probabilistic imputation~\cite{vae}. MIWAE~\cite{miwae} employs importance-weighted autoencoders to tighten the bound for improved log-likeliho\-od estimation, while HI-VAE~\cite{hivae} handles mixed-type data through heterogeneous likelihoods. However, these methods tend to produce over-smoothed estimates due to inference suboptimality~\cite{vae-}, and the restrictive Gaussian prior limits their capacity to capture complex multi-modal distributions~\cite{vae-2}.

    \textbf{Graph and Transformer-based methods.} GRAPE~\cite{grape} and IGRM~\cite{igrm} represent the data matrix as a bipartite graph and recover missing entries via message passing between samples and feature nodes.    
    ReMasker~\cite{remasker} adapts the masked autoencoding paradigm and learns to reconstruct randomly masked features via Transformer-based self-attention.
    These methods may incur overhead from graph construction and quadratic attention complexity~\cite{transformer-}.

    \textbf{Optimal Transport-based methods.} OTImpute~\cite{otimpute} aligns distributions by minimizing Sinkhorn divergence~\cite{sinkhorn} between observed and imputed data, while TDM~\cite{tdm} further applies invertible transformations before distribution matching.
    Despite their sound theoretical foundation, these methods typically require solving expensive optimization subroutines for divergence computation, leading to substantial training costs~\cite{ot-}.
    
    \textbf{Diffusion-based methods.} TabCSDI~\cite{tabcsdi} models data generation as a reverse stochastic process, gradually denoising random noise conditioned on observed values. DiffPuter~\cite{diffputer} integrates denoising within an EM framework, NewImp~\cite{newimp} reformulates imputation via gradient flow, and SimpDM~\cite{simpdm} stabilizes accuracy through self-supervised regularization. Although achieving superior fidelity, diffusion models suffer from high inference latency due to the need for hundreds of iterative sampling steps~\cite{diffusion-,diffusionsurvey}.

    \subsection{Unsupervised Pattern Analysis Techniques}     
    Early approaches rely on parametric or linear assumptions, and thus struggle to capture complex non-linear dependencies. Accordingly, non-parametric approaches, such as $K$-Means~\cite{kmeans2}, leverage group-level structure but are vulnerable to seed initialization and data heterogeneity, i.e., a mixed dataset composed of both numerical and categorical attributes. 

    In recent years, significant progress has been made in robust mixed data clustering. Recent advancements have developed novel distance metrics~\cite{zhang2022graph}, hierarchical merging~\cite{cai2024robust}, multi-granular distributions~\cite{zhang2025adaptive}, and homogeneous concept spaces~\cite{zhang2022het2hom} to accurately partition the complex mixed data~\cite{missforest,zhao2022heterogeneous}. Concurrently, exploring value orders~\cite{zhang2020ordinal}, developing order-inspired metrics~\cite{zhang2026categorical}, learning order forests~\cite{zhao2024learning}, as well as cluster-customized category relationships~\cite{zhao2026break} have proven highly effective for categorical data clustering.

\section{Proposed Method} \label{sec:method}
\begin{figure}[t]
    \centering
    \includegraphics[width=1\textwidth]{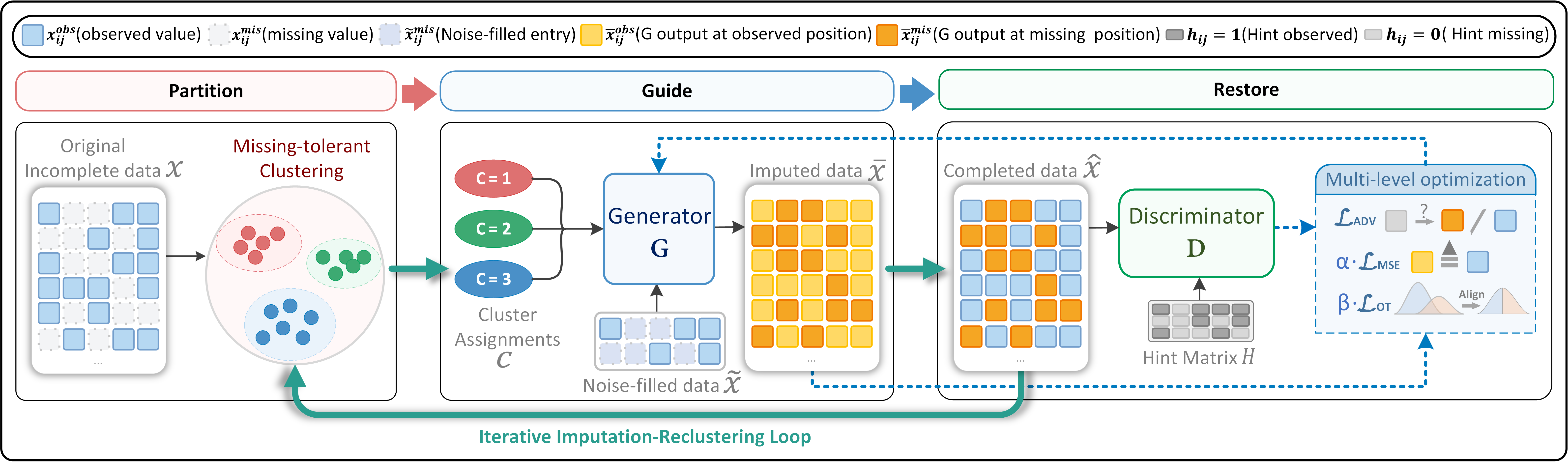}
        \caption{The overall architecture of CAGI framework following a ``Partition-Guide-Restore'' design: the missing-tolerant clustering module first assigns the incomplete data into clusters and treats the labels as conditioning signals for generator $G$. Then $G$ imputes all positions to produce $\bar{x}$, and the completed data $\hat{x}$ retains original observed values and replaces missing entries with $\bar{x}_{ij}^{mis}$, which is fed to discriminator $D$ alongside the hint matrix. $D$ and $G$ engage in adversarial training where $D$ distinguishes observed from imputed entries while $G$ deceives $D$. An iterative imputation-reclustering process updates the assignments until convergence, yielding $\hat{x}$ as the final imputed output.} \label{fig:framework}
    \end{figure}

    To address the challenge of imputing missing values in data with complex latent structures, we propose the CAGI framework (Figure~\ref{fig:framework}). 
    CAGI follows a ``Partition–Guide–Restore'' principle, and its imputation  architecture consists of three key components: 
    (1) a missing‑tolerant clustering module that discovers subgroups from incomplete data,  
    (2) a cluster‑conditioned generator that performs subgroup‑aware imputation, and 
    (3) an adversarial mechanism that enforces element‑wise realism. 
    These components are integrated through a multi‑level learning objective and a progressive alternating training procedure. 
    In the following, we first introduce the problem formulation, then detail the imputation architecture, learning strategy, and alternating training loop.

    \subsection{Problem Formulation}     
    Let $\mathbf{X} \in \mathbb{R}^{n \times d}$ denote a dataset with $n$ samples and $d$ features, partially observed under a binary mask matrix $\mathbf{M} \in \{0,1\}^{n \times d}$, where $m_{ij}=1$ indicates an observed entry. Each sample $\mathbf{x}_i$ is decomposed into observed components $\mathbf{x}_i^{obs}$ and missing components $\mathbf{x}_i^{mis}$. The imputation goal is to estimate $\mathbf{x}_i^{mis}$ by modeling $P(\mathbf{x}^{mis} \mid \mathbf{x}^{obs})$.
    CAGI leverages clustering to uncover latent subgroups $\{c_1,\dots,c_K\}$ in real-world data. By incorporating the cluster assignment into the imputation process, CAGI targets the cluster-conditional distribution $P(\mathbf{x}^{mis} \mid \mathbf{x}^{obs}, c)$. Within a GAN~\cite{gan} framework, the generator's raw output is:
    \begin{equation}
      \bar{\mathbf{x}} = G(\tilde{\mathbf{x}},\, \mathbf{m},\, c),
    \end{equation}
    where $\tilde{\mathbf{x}}$ fills missing entries with noise, $\mathbf{m}$ is the mask, and $c$ is the cluster assignment. The completed sample merges generated and observed values:
    \begin{equation}
      \hat{\mathbf{x}} = \mathbf{m} \odot \mathbf{x} + (1 - \mathbf{m}) \odot \bar{\mathbf{x}},
    \end{equation}
    where $\odot$ denotes element-wise multiplication. By conditioning on $c$, the model narrows the generation scope to the relevant subgroup, thereby ensuring that the imputed values reflect the characteristics of the specific underlying subgroup.
    
    \subsection{Cluster-Aware Generative Imputation} \label{sec:method_design}
    To resolve the circular dependency discussed in Section~\ref{sec:intro}, CAGI unifies clustering and imputation into an alternating optimization process. The architecture is detailed below, organized around the ``Partition-Guide-Restore'' strategy.

    \subsubsection{Partition: Missing-Tolerant Clustering.}
    Standard clustering algorithms (e.g., $K$-Means) cannot operate on incomplete data, as the standard Euclidean distance cannot be computed when features are partially missing. 
    To initiate the process, we employ a missing-tolerant clustering mechanism that computes partial distances based solely on shared observed features.
    This strategy follows the well-established practice of partial-distance clustering~\cite{clusterformiss1,clusterformiss2}, which allows standard clustering algorithms to operate directly on incomplete data without requiring prior imputation.
    For any two samples $\mathbf{x}_i$ and $\mathbf{x}_j$, the set of commonly observed features is:
    \begin{equation}
        \mathcal{S}_{ij} = \{k : m_{ik}=1 \wedge m_{jk}=1\}.
    \end{equation}
    The partial distance between two samples is computed as:
    \begin{equation}
        d(\mathbf{x}_i, \mathbf{x}_j) = \sqrt{\frac{d}{|\mathcal{S}_{ij}|} \sum_{k \in \mathcal{S}_{ij}} (x_{ik} - x_{jk})^2}, \label{eq:pd}
    \end{equation}
    where the scaling factor $\frac{d}{|\mathcal{S}_{ij}|}$ normalizes the partial sum to approximate the full-dimensional Euclidean distance, ensuring that sample pairs with different numbers of co-observed features remain comparable. When $\mathcal{S}_{ij}= \emptyset$, the distance is set to infinity and the sample is assigned to the nearest cluster with which it shares observed features. Based on this distance metric, we assign each sample to a cluster $c_i \in \{1, \dots, K\}$, providing the initial subgroup guidance for the generator. Note that $K$ need not match the true number of subgroups. Rather, the purpose of clustering is to constrain the generator within localized regions of the feature space, reducing the risk of cross-subgroup imputation illustrated in Figure~\ref{fig:motivation}. 
    
    For purely numerical datasets, standard $K$-Means is used with the partial distance metric defined in Eq.~(\ref{eq:pd}). 
    For datasets containing categorical features, we employ $K$-Prototypes~\cite{kprototype}, which combines Euclidean distance for numerical features with the simple matching distance for categorical features.
    Specifically, for two samples $\mathbf{x}_i$ and $\mathbf{x}_j$, the combined distance is defined as:
    \begin{equation}
    d_{\text{mix}}(\mathbf{x}_i, \mathbf{x}_j) = \sqrt{\frac{d_{num}}{|\mathcal{S}_{ij}^{num}|} \sum_{k \in \mathcal{S}_{ij}^{num}} (x_{ik} - x_{jk})^2} + \gamma \cdot \frac{d_{cat}}{|\mathcal{S}_{ij}^{cat}|} \sum_{k \in \mathcal{S}_{ij}^{cat}} \delta_{ij}^{k},
    \end{equation}
    \begin{equation}
    \delta_{ij}^{k} = 
    \begin{cases}
        0, & \text{if } x_{ik} = x_{jk}, \\
        1, & \text{if } x_{ik} \neq x_{jk}.
    \end{cases}
    \end{equation}
    $d_{num}$ and $d_{cat}$ denote the total number of numerical and categorical features, $\mathcal{S}_{ij}^{num}$ and $\mathcal{S}_{ij}^{cat}$ are the sets of co-observed numerical and categorical features between samples $\mathbf{x}_i$ and $\mathbf{x}_j$, and $\gamma$ is a balancing parameter that controls the relative contribution of categorical dissimilarity.
  
    \subsubsection{Guide: Cluster-Conditioned Generation.}
    With the initial cluster assignments established, the generator $G$ receives a composite input consisting of the noise-filled data $\tilde{\mathbf{x}}_i$, the mask $\mathbf{m}_i$, and the cluster assignment $c_i$.
    Missing positions are initialized with feature-wise noise $\mathbf{z}_i$ whose $j$-th entry is sampled as $z_{ij} \sim \mathcal{N}(\mu_j, (0.1 \cdot \sigma_j)^2)$:
    \begin{equation}
        \tilde{\mathbf{x}}_i = \mathbf{m}_i \odot \mathbf{x}_i + (1 - \mathbf{m}_i) \odot \mathbf{z}_i.
    \end{equation}
    The mean $\mu_{j}$ anchors the initialization at the center of the observed distribution, while the small variance factor $0.1$ keeps the initial perturbation mild, preventing the noise from dominating the early generator output. This initialization serves only as a starting point: as training proceeds, the generator progressively overwrites the initial values through the iterative loop (Section~\ref{sec:training_loop}), so the final performance is not tightly coupled to this specific choice.
    
    The raw generator output $\bar{\mathbf{x}}_i = G(\tilde{\mathbf{x}}_i, \mathbf{m}_i, c_i)$ is then combined with observed values to produce the completed sample:
    \begin{equation}
        \hat{\mathbf{x}}_i = \mathbf{m}_i \odot \mathbf{x}_i + (1 - \mathbf{m}_i) \odot \bar{\mathbf{x}}_i.
    \end{equation}
    By conditioning on $c_i$, $G$ learns a subgroup‑specific mapping $G(\cdot \mid c{=}k)$, narrowing the generative search space and reducing generation uncertainty.

    \subsubsection{Restore: Adversarial Imputation.}
    The restoration quality is enforced by an adversarial mechanism adapted from GAIN~\cite{gain}.
    A discriminator $D$ evaluates each completed sample $\hat{\mathbf{x}}_i$ in an element‑wise manner, outputting the probability that each feature corresponds to an observed entry. Formally, $D: \mathbb{R}^d \times \{0,1\}^d \to [0, 1]^d$, where the $j$-th output $D(\hat{\mathbf{x}}, \mathbf{h})_j$ estimates the probability that the $j$-th feature of $\hat{\mathbf{x}}$ is observed.
    To prevent the discriminator from trivially memorizing the mask pattern, a hint mechanism is incorporated. The hint matrix $\mathbf{h}_i \in \{0, 1\}^d$ reveals partial information about which positions are observed:   
    \begin{equation}
        \mathbf{h}_i = \mathbf{b}_i \odot \mathbf{m}_i,
    \end{equation}
    where $\mathbf{b}_i \in \{0, 1\}^d$ is sampled element-wise from $\mathrm{Bernoulli}(p_h)$. Entries with $h_{ij} = 1$ reveal observed positions, whereas entries with $h_{ij} = 0$ leave the status ambiguous.
    Through this adversarial structure, the generator is trained to produce imputations that are indistinguishable from real values at the element level, promoting convergence toward the true conditional distribution.

    \subsection{Multi-Level Learning Objective} \label{sec:method_obj}
    A potential challenge of cluster-conditioned generation is distributional fragmentation, particularly when early cluster assignments are inaccurate. CAGI addresses this through a composite objective operating at two complementary levels: instance-level fidelity and distribution-level consistency.
    
    \textbf{Instance-Level Fidelity.} 
    CAGI enforces element-wise accuracy through adversarial learning and reconstruction consistency.
    The discriminator $D$ is trained to maximize correct identification of observed versus imputed entries:
    \begin{equation} \label{eq:L_D}
        \mathcal{L}_D = - \mathbb{E}_{\hat{\mathbf{x}}, \mathbf{m}, \mathbf{h}} \Big[ \mathbf{m} \odot \log D(\hat{\mathbf{x}}, \mathbf{h}) + (1 - \mathbf{m}) \odot \log \big(1 - D(\hat{\mathbf{x}}, \mathbf{h})\big) \Big].
    \end{equation}
    The generator $G$ aims to deceive $D$ specifically on the missing components:
    \begin{equation} \label{eq:L_ADV}
        \mathcal{L}_{ADV} = - \mathbb{E}_{\hat{\mathbf{x}}, \mathbf{m}, \mathbf{h}} \Big[ (1 - \mathbf{m}) \odot \log D(\hat{\mathbf{x}}, \mathbf{h}) \Big].
    \end{equation}
    A reconstruction constraint on observed entries penalizes the generator for distorting positions where the ground truth is known, encouraging it to learn an identity mapping on observed features:
    \begin{equation} \label{eq:L_MSE}
        \mathcal{L}_{MSE} = \frac{1}{|\mathcal{O}|} \sum_{(i,j) \in \mathcal{O}} (x_{ij} - \bar{x}_{ij})^2,
    \end{equation}
    where $\mathcal{O} = \{(i,j): m_{ij} = 1\}$ denotes the set of observed entries, and $\bar{x}_{ij}$ is the raw generator output before combining with observed values. 

    \textbf{Distribution-Level Consistency.}   
    While cluster conditioning guides the generator toward subgroup‑specific subspaces, adversarial training on incomplete data may introduce distributional fragmentation. To promote a coherent manifold structure, a distribution‑level regularizer based on Optimal Transport (OT) is incorporated.
    Specifically, the Sinkhorn divergence~\cite{sinkhorn} is adopted to enforce self‑consistency within the generated distribution. For a mini-batch of completed samples $\hat{\mathcal{B}} = \{\hat{\mathbf{x}}_1, \ldots, \hat{\mathbf{x}}_b\}$, CAGI randomly partitions them into two disjoint subsets $\hat{\mathcal{B}}_1$ and $\hat{\mathcal{B}}_2$ of equal size, then minimizes their transport cost:
    \begin{equation} \label{eq:L_OT}
        \mathcal{L}_{OT} = Sinkhorn_{\epsilon}(\hat{\mathcal{B}}_1, \hat{\mathcal{B}}_2),
    \end{equation}
    where $Sinkhorn_{\epsilon}$ denotes the optimal transport cost with regularization parameter $\epsilon$.
    The entropy regularization smooths the transport plan so that the divergence reflects distributional structure rather than individual point positions, yielding reliable estimates even on mini-batch-sized subsets~\cite{sinkhorn}. 
    The underlying intuition is that two random subsets drawn from a stable distribution should be statistically indistinguishable. By minimizing the divergence between random partitions, CAGI penalizes geometric irregularities and encourages a compact manifold, complementing the element-wise reconstruction objective.
    
    \textbf{Composite Objective.}
    The final generator objective balances instance-level fidelity and distribution-level consistency:
    \begin{equation}\label{eq:L_G}
        \mathcal{L}_G = \mathcal{L}_{ADV} + \alpha \cdot \mathcal{L}_{MSE} + \beta \cdot \mathcal{L}_{OT},
    \end{equation}
    where $\alpha$ controls the weight of the reconstruction constraint and $\beta$ governs the strength of the distributional regularization. 
    The three loss components serve complementary roles: $\mathcal{L}_{MSE}$ anchors instance-level fidelity by leveraging observed ground truth, $\mathcal{L}_{ADV}$ promotes statistical realism through adversarial signals, and $\mathcal{L}_{OT}$ enforces global distributional coherence to prevent fragmentation induced by cluster conditioning. 
    Overall, the complete training procedure alternates between updating $D$ via Eq.~(\ref{eq:L_D}) and updating $G$ via Eq.~(\ref{eq:L_G}).

    \subsection{Progressive Alternating Training} \label{sec:training_loop}
    A core proposition is the dynamic \emph{imputation-clustering loop} that treats cluster assignments as continuously updated latent variables rather than fixed information.   
    Every $T$ iterations, the current generator $G^{(t)}$ completes the training set to obtain $\hat{\mathbf{X}}$, while standard $K$-Means is then applied to $\hat{\mathbf{X}}$ to update assignments $\mathcal{C}^{(t)}$ based on the currently available features, which can be summarized as:   
    
    \noindent Step 1 - \textbf{Impute}. Generate completed data using the current model:
        \begin{equation}
            \hat{\mathbf{x}}_i^{(t)} = \mathbf{m}_i \odot \mathbf{x}_i + (1 - \mathbf{m}_i) \odot G^{(t)}(\tilde{\mathbf{x}}_i, \mathbf{m}_i, c_i^{(t)});
        \end{equation}        
    \noindent Step 2 - \textbf{Reclustering}. Re-apply clustering on the completed data:
        \begin{equation}
            \mathcal{C}^{(t+1)} = Clustering(\{\hat{\mathbf{x}}_1^{(t)}, \ldots, \hat{\mathbf{x}}_n^{(t)}\}, K).
        \end{equation}
    This feedback loop creates mutually reinforcing trends: improved cluster guidance reduces generative uncertainty, while refined imputations correct early-stage assignment errors. Crucially, the $\mathcal{L}_{MSE}$ term anchors this process to observed ground-truth signals, preventing divergence.
    At inference time, a new incomplete sample is assigned to its nearest cluster via the missing-tolerant distance metric (Eq.~\ref{eq:pd}), and then imputed by the trained generator conditioned on the identified cluster assignment. 
    
    It is also noteworthy that the per-epoch training cost of CAGI scales linearly with sample size $n$, achieving a favorable trade-off between imputation quality and computational efficiency.

\section{Experiments} \label{sec:exp}
    This section evaluates CAGI from four perspectives:
    (i) imputation fidelity against 15 state-of-the-art baselines across 14 datasets, 
    (ii) the contribution of each design component through ablation,
    (iii) sensitivity to key hyperparameters and scalability, 
    (iv) downstream utility on classification and clustering tasks.

    \subsection{Experimental Settings}
    \subsubsection{Datasets.} 
    The evaluation uses 14 public real-world datasets. We consider six numerical datasets (\textit{spam}, \textit{breast}, \textit{wine}, \textit{yeast}, \textit{blood}, \textit{california}), three categorical datasets (\textit{car}, \textit{mushroom}, \textit{letter}), and five mixed-type datasets (\textit{credit}, \textit{adult}, \textit{shoppers}, \textit{heart}, \textit{news}). 
    Numerical features are normalized to $[0, 1]$ via Min-Max scaling and categorical features are transformed via the Analog Bits encoding scheme~\cite{analog}. Compared to one-hot encoding, Analog Bits yields a more compact representation that scales logarithmically with the number of categories, particularly beneficial for high-cardinality features.

    \subsubsection{Baselines.} CAGI is compared against 15 representative imputation methods spanning statistical, machine learning, and deep learning paradigms: Mean/Mode\allowbreak{} ~\cite{mean}, KNNImpute~\cite{knn}, MICE~\cite{mice}, MissForest~\cite{missforest}, GAIN~\cite{gain}, MIWAE~\cite{miwae}, MIRACLE~\cite{miracle}, OTImpute~\cite{otimpute}, TDM~\cite{tdm}, GCImpute~\cite{gc}, TabCSDI~\cite{tabcsdi}, GRA\-PE~\cite{grape}, ReMasker~\cite{remasker}, IGRM~\cite{igrm}, and DiffPuter~\cite{diffputer}. 

    \subsubsection{Metrics.} 
    Root Mean Square Error (RMSE) is adopted for numerical attributes and Proportion of Falsely Classified entries (PFC)~\cite{missforest} for categorical attributes, where imputed continuous values are decoded back to discrete categories before evaluation. Lower values indicate better performance for both. Downstream utility is assessed through two tasks: (1) \textit{Classification}, where a Logistic Regression classifier is trained on imputed data and evaluated by Area Under the ROC Curve (AUROC), and (2) \textit{Clustering}, where $K$-Means is applied to imputed data with the number of clusters matching the ground-truth class count, and Adjusted Rand Index (ARI) is reported.
    Following prior work~\cite{gain,grape,remasker,diffputer}, missing values are generated under the multivariate MCAR setting, where missingness is introduced independently across all features and samples.
    Unless otherwise stated, the missing rate is set to 0.6 to evaluate methods under challenging conditions.
    Each experiment is repeated five times with 5-fold cross-validation, and the average performance is reported. In each repetition, a new MCAR mask is independently generated, and imputation models are fitted on the training fold only, with imputation quality evaluated on the test fold.

    \begin{table}[t]
        \centering
        \caption{Comparison of imputation performance on numerical datasets (measured by RMSE). The \colorbox{bestc}{best} and \colorbox{secondc}{second-best} results are highlighted for each dataset.}
        \label{tab:imp_rmse}         
            \begin{tabular}{l|cccccc}
                \toprule
                Method & \textbf{spam} & \textbf{breast} & \textbf{wine} & \textbf{yeast} & \textbf{blood} & \textbf{california}\\
                \midrule                
                Mean/Mode  & $.0559_{\pm .0025}$ & $.1459 _{\pm .0078}$ & $.1059 _{\pm .0016}$ & $.1239 _{\pm .0044}$ & $.1660 _{\pm .0153}$ & $.1628 _{\pm .0009}$ \\
                KNNImpute  & $.0634 _{\pm .0026}$ & $.1182 _{\pm .0046}$ & $.1106 _{\pm .0021}$ & $.1283 _{\pm .0047}$ & $.1451 _{\pm .0133}$ & $.1645 _{\pm .0011}$ \\
                MICE       & $.0729 _{\pm .0011}$ & $.1038 _{\pm .0044}$ & $.1335 _{\pm .0021}$ & $.1638 _{\pm .0062}$ & $.1884 _{\pm .0145}$ & $.1945 _{\pm .0009}$ \\
                MissForest & $.0604 _{\pm .0037}$ & $.0982 _{\pm .0032}$ & $.1087 _{\pm .0023}$ & $.1231 _{\pm .0048}$ & \cellcolor{bestc}$.1270 _{\pm .0044}$ & $.1531 _{\pm .0040}$ \\
                GAIN       & $.0743 _{\pm .0212}$ & $.1239 _{\pm .0189}$ & $.1613 _{\pm .0206}$ & $.1369 _{\pm .0123}$ & $.1553 _{\pm .0217}$ & $.2120 _{\pm .0310}$ \\
                MIWAE      &$.0561 _{\pm .0017}$ &\cellcolor{secondc}$.0916 _{\pm .0028}$ & $.1078 _{\pm .0009}$ & $.1298 _{\pm .0057}$ & $.1349 _{\pm .0110}$  & \cellcolor{bestc}$.1410_{\pm .0016}$\\
                MIRACLE    & $.0560 _{\pm .0026}$ & $.1484 _{\pm .0147}$ & $.1116 _{\pm .0022}$ & $.1236 _{\pm .0049}$ & $.1445 _{\pm .0149}$ & $.1632 _{\pm .0047}$ \\
                OTImpute   & $.0654 _{\pm .0014}$ & $.0951 _{\pm .0025}$ & $.1129 _{\pm .0023}$ & $.1454 _{\pm .0048}$ & $.1751 _{\pm .0101}$ & $.1796 _{\pm .0009}$ \\
                TDM        & $.1073 _{\pm .0008}$ & $.1651 _{\pm .0063}$ & $.1320 _{\pm .0018}$ & $.2023 _{\pm .0283}$ & $.1969 _{\pm .0033}$ & $.2121 _{\pm .0205}$ \\
                GCImpute   & $.1442 _{\pm .0017}$ & $.0975 _{\pm .0055}$ & $.1045 _{\pm .0016}$ & $.1340 _{\pm .0189}$ & $.1508 _{\pm .0135}$ & $.1442 _{\pm .0014}$ \\
                TabCSDI    & $.0704 _{\pm .0046}$ & $.1103 _{\pm .0056}$ & $.1231 _{\pm .0021}$ & $.1452 _{\pm .0017}$ & $.1832 _{\pm .0115}$ & $.1551 _{\pm .0011}$ \\
                GRAPE      & $.0604 _{\pm .0022}$ & $.1062 _{\pm .0065}$ & $.1047 _{\pm .0017}$ & $.1205 _{\pm .0057}$ &\cellcolor{secondc}$.1325 _{\pm .0125}$ & $.1441 _{\pm .0027}$ \\
                ReMasker   &\cellcolor{secondc} $.0546 _{\pm .0020}$ & $.1227 _{\pm .0037}$ & $.1088 _{\pm .0017}$ & $.1396 _{\pm .0035}$ & $.1620 _{\pm .0020}$ & $.1455 _{\pm .0028}$\\
                IGRM       & $.0571 _{\pm .0019}$ & $.1000 _{\pm .0073}$ & $.1111 _{\pm .0016}$ & $.1243 _{\pm .0047}$ & $.1557 _{\pm .0146}$ & $.1578 _{\pm .0015}$ \\
                DiffPuter  & $.0564 _{\pm .0027}$ & $.1032 _{\pm .0031}$ & \cellcolor{secondc}$.1029 _{\pm .0023}$ &\cellcolor{secondc}$.1195 _{\pm .0025}$ & $.1421 _{\pm .0161}$ & \cellcolor{secondc} $.1415 _{\pm .0006}$ \\
                CAGI (ours)      & \cellcolor{bestc}$.0532 _{\pm .0016}$ &\cellcolor{bestc} $.0884 _{\pm .0053}$ & \cellcolor{bestc}$.0997 _{\pm .0014}$ & \cellcolor{bestc}$.1173 _{\pm .0063}$ & $.1539 _{\pm .0144}$ & $.1489 _{\pm .0014}$ \\
                \bottomrule
            \end{tabular}
    \end{table}

    \begin{table}[t]
        \centering
        \caption{Comparison of imputation performance on categorical datasets (measured by PFC). The \colorbox{bestc}{best} and \colorbox{secondc}{second-best} results are highlighted for each dataset.}
        \label{tab:imp_pfc}         
            \begin{tabular}{l|ccc}
                \toprule
                Method & \textbf{car} & \textbf{mushroom} & \textbf{letter} \\
                \midrule  
                Mean/Mode  & \cellcolor{secondc}$.6582 _{\pm .0212}$ & $.4101 _{\pm .0070}$ & $.7690 _{\pm .0035}$ \\
                KNNImpute  & $.6664 _{\pm .0097}$ & $.3399 _{\pm .0037}$ & $.7898 _{\pm .0019}$\\
                MICE       & $.6684 _{\pm .0128}$ & $.3369 _{\pm .0068}$ & $.8245 _{\pm .0022}$ \\
                MissForest & $.6887 _{\pm .0136}$ & $.2774 _{\pm .0094}$ & $.7742 _{\pm .0134}$ \\
                GAIN       & $.6665 _{\pm .0121}$ & $.4449 _{\pm .0228}$ & $.8487 _{\pm .0164}$ \\
                MIWAE      & $.6749 _{\pm .0195}$ & \cellcolor{secondc}$.2720 _{\pm .0080}$ & $.7663 _{\pm .0024}$ \\
                MIRACLE    & $.6977 _{\pm .0206}$ & $.3161 _{\pm .0259}$ & $.8507 _{\pm .0106}$ \\
                OTImpute   & $.6814 _{\pm .0128}$ & $.3029 _{\pm .0034}$ & $.7489 _{\pm .0019}$ \\
                TDM        & $.6624 _{\pm .0119}$ & $.5135 _{\pm .0099}$ & $.8720 _{\pm .0032}$ \\
                GCImpute   & $.6611 _{\pm .0104}$ & $.3854 _{\pm .0294}$ & \cellcolor{secondc}$.7438 _{\pm .0026}$ \\
                TabCSDI    & $.6675 _{\pm .0140}$ & $.3130 _{\pm .0085}$ & $.7770 _{\pm .0039}$ \\
                GRAPE      & $.6626 _{\pm .0141}$ & $.3898 _{\pm .0303}$ & $.7718 _{\pm .0057}$ \\
                ReMasker   & $.6986 _{\pm .0129}$ & $.2859 _{\pm .0054}$ & $.7524 _{\pm .0032}$ \\
                IGRM       & $.6747 _{\pm .0160}$ & $.3454 _{\pm .0082}$ & $.7793 _{\pm .0040}$ \\
                DiffPuter  & $.6684 _{\pm .0155}$ & $.2952 _{\pm .0098}$ & $.7521 _{\pm .0037}$ \\
                CAGI (ours)& \cellcolor{bestc} $.6539 _{\pm .0126}$ & \cellcolor{bestc}$.2532 _{\pm .0032}$ &\cellcolor{bestc} $.7333 _{\pm .0027}$ \\
                \bottomrule
            \end{tabular}
    \end{table}
    
    \begin{figure}[t] 
        \centering 
        \includegraphics[width=\textwidth]{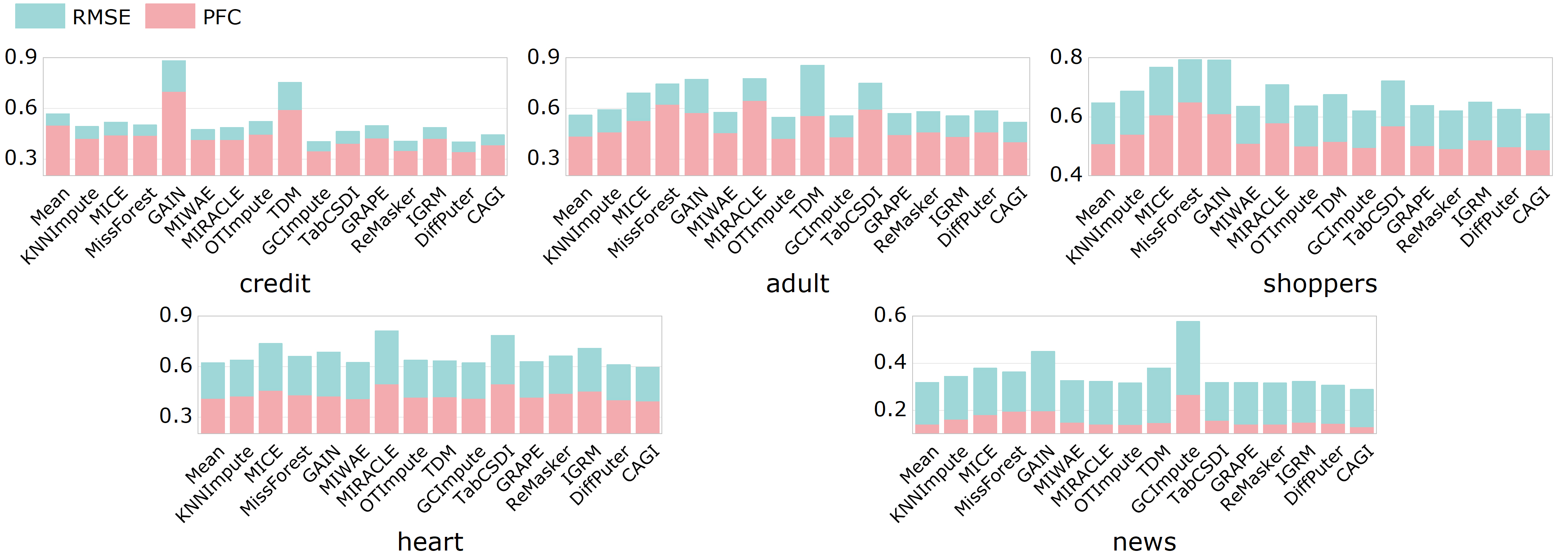}
        \caption{Imputation performance comparison on mixed-type datasets. Each stacked bar shows RMSE (top) and PFC (bottom) simultaneously.} \label{fig:imp_mix}
    \end{figure}

    \subsection{Imputation Performance} 
    Tables~\ref{tab:imp_rmse},~\ref{tab:imp_pfc} and Figure~\ref{fig:imp_mix} present imputation performance on single-type and mixed-type datasets, respectively.
    CAGI ranks first on 7 out of 9 single-type datasets and achieves the lowest combined error on 4 out of 5 mixed-type datasets. 
    Although no single method dominates across every dataset due to varying data characteristics, CAGI's consistent superiority across diverse benchmarks supports the central hypothesis: by partitioning the generative search space via cluster conditioning, the imputation model approximates a collection of simpler subgroup-specific distributions rather than a single complex global one, yielding higher instance-level fidelity.
    CAGI also demonstrates strong performance on mixed data, as the subgroup-specific co-occurrence patterns between numerical and categorical features are captured by the combined distance metric.

    \subsection{Ablation Study} \label{sec:ab}
        \begin{table}[t]
        \centering
        \caption{Ablation study on key components and loss functions.}
        \label{tab:abla} 
            \begin{tabular}{l c c c c}
                \toprule
                \multirow{2.5}{*}{\textbf{Method}} & \textbf{spam} & \textbf{letter} & \multicolumn{2}{c}{\textbf{adult}} \\
                \cmidrule(lr){2-2} \cmidrule(lr){3-3} \cmidrule(lr){4-5}
                & \footnotesize RMSE  & \footnotesize PFC & \footnotesize RMSE  & \footnotesize PFC  \\
                \midrule
                
            Full Model               & $\mathbf{.0532} _{\pm .0016}$     & $\mathbf{.7333} _{\pm .0027}$    & $\mathbf{.1177}_{\pm .0011}$ & $ \mathbf{.4006}_{\pm .0040}$\\
                
            \midrule
            \multicolumn{5}{l}{\textcolor{gray}{\textbullet\ \textit{Impact of Key Components}}} \\
                \hspace{1em} w/o \textit{Clustering}  & $.0598_{\pm .0014}$     & $.7548_{\pm  .0028}$     & $.1239_{\pm .0013}$  & $.4103_{\pm .0034}$ \\
                \hspace{1em} w/o \textit{Updating}    & $.0580_{\pm .0029}$     & $.7369_{\pm .0036}$     & $.1193_{\pm .0013}$  & $.4022_{\pm .0035}$  \\
                    
            \specialrule{\lightrulewidth}{3pt}{1pt} 
            \specialrule{\lightrulewidth}{0pt}{3pt}
            \multicolumn{5}{l}{\textcolor{gray}{\textbullet\ \textit{Impact of Loss Functions}}} \\ 
        
                \hspace{1em} w/o $\mathcal{L}_{\text{MSE}}$  & $.1224_{\pm .0161}$  & $.7525_{\pm  .0047}$  & $.1447_{\pm .0260}$ & $.4060_{\pm .0032}$ \\
                
                \hspace{1em} w/o $\mathcal{L}_{\text{ADV}}$  & $.0591_{\pm .0016}$  & $.7351 _{\pm  .0027}$  & $.1178_{\pm .0013}$ & $.4015_{\pm .0034}$ \\

                \hspace{1em} w/o $\mathcal{L}_{\text{OT}}$   & $.0586_{\pm .0014}$  & $.7731_{\pm  .0030}$  & $.1187_{\pm  .0013}$ & $.4242_{\pm .0050}$ \\
                
                \bottomrule
            \end{tabular}
        
    \end{table}
    Table~\ref{tab:abla} validates the design of CAGI by removing key modules and loss terms.
    
    \textbf{Impact of Key Components.} 
    Removing cluster conditioning (``w/o \textit{Clustering}'') leads to degradation across all datasets, with the largest drop on \textit{spam}. Without subgroup priors, the generator approximates the entire data distribution as a single manifold and loses the ability to capture subgroup-specific patterns. This directly validates CAGI's core design premise. 
    Disabling periodic reclustering (``w/o \textit{Updating}'') reduces CAGI to a fixed cluster-then-impute pipeline. The consistent degradation confirms the value of iterative co-refinement over static assignments, even when initial clusters are imperfect.
   
    \textbf{Impact of Loss Functions.} 
    The reconstruction loss $\mathcal{L}_{\text{MSE}}$ proves to be the most critical component: its removal causes RMSE to increase dramatically, confirming that $\mathcal{L}_{\text{MSE}}$ serves as the primary anchor for instance-level fidelity.
    The Sinkhorn regularization $\mathcal{L}_{\text{OT}}$ has a strong impact on categorical data: removing it leads to a notable degradation in PFC. This aligns with the design rationale that distribution-level consistency is particularly important for categorical features, where small distributional shifts can flip discrete category assignments.
    Removing $\mathcal{L}_{\text{ADV}}$ also degrades performance, confirming that adversarial training further improves the statistical quality and training stability of the imputation.

    \begin{figure}[t]
    \centering 
        \begin{minipage}{0.32\textwidth} 
        \centering 
        \includegraphics[width=\textwidth]{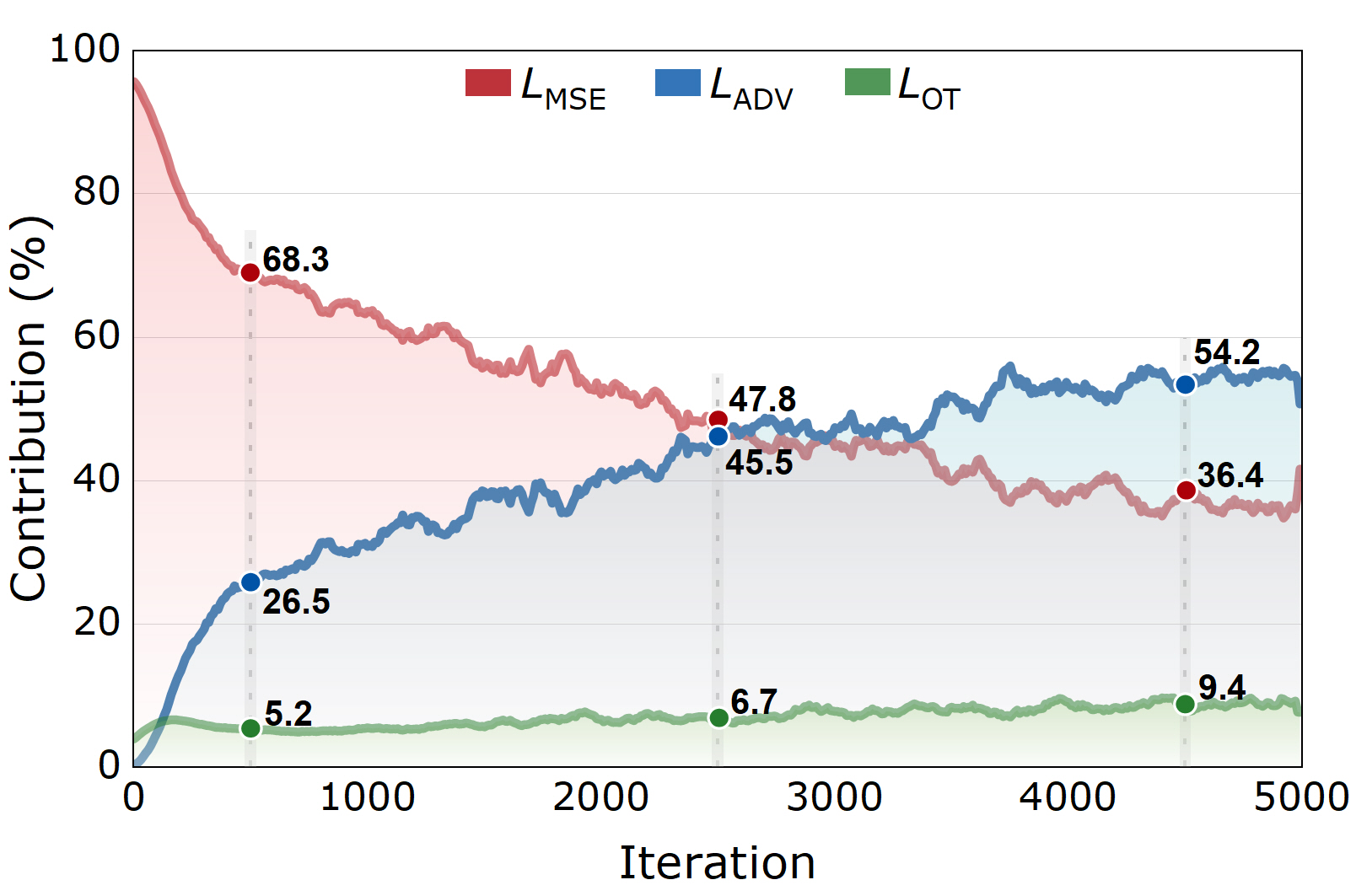} 
        \caption{The contribution ratio of each loss component ($\mathcal{L}_{\text{MSE}}$, $\mathcal{L}_{\text{ADV}}$, and $\mathcal{L}_{\text{OT}}$).} \label{fig:training_loss}
        \end{minipage} 
        \hfill 
        \begin{minipage}{0.64\textwidth}
        \centering 
        \includegraphics[width=\textwidth]{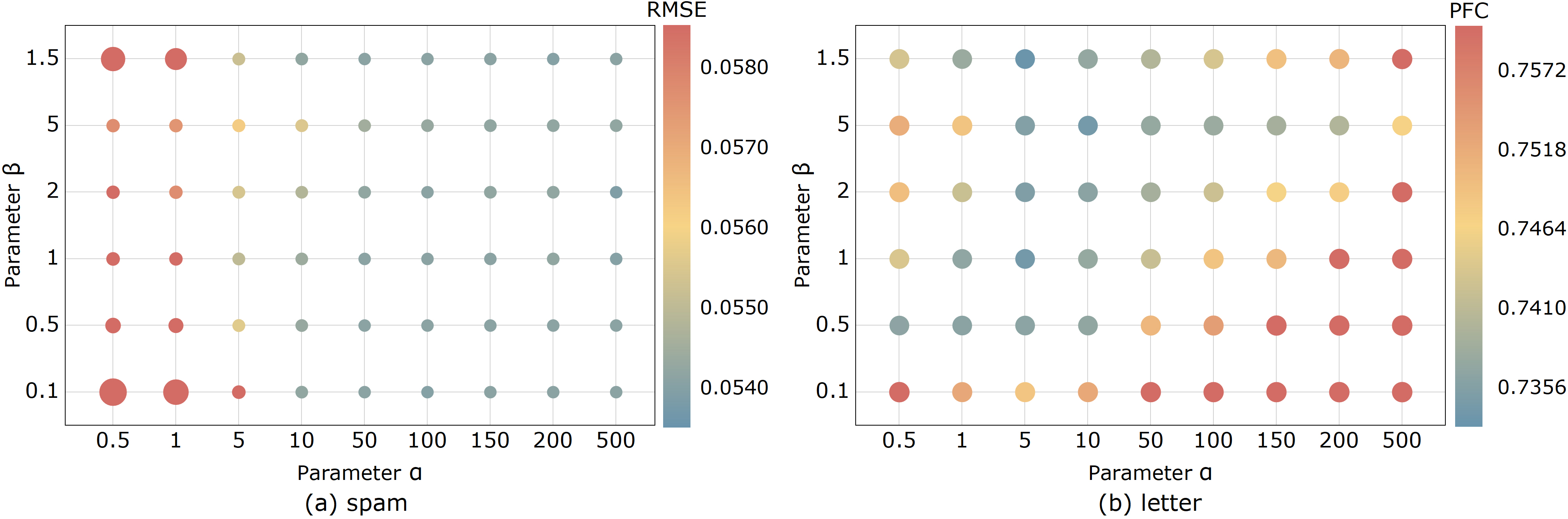} 
        \caption{Sensitivity to loss weights $\alpha$ ($\mathcal{L}_{\text{MSE}}$) and $\beta$ ($\mathcal{L}_{\text{OT}}$) on datasets \textit{spam} (a) and \textit{letter} (b). Metric values are indicated by bubble size and color.} \label{fig:training_ab}
        \end{minipage} 
    \end{figure}

    \textbf{Training Loss Analysis.} To further understand how the multi-level objectives interact during optimization, Figure~\ref{fig:training_loss} visualizes the relative contribution of each loss term over training on the \textit{spam} dataset.
    The optimization naturally progresses through three phases: 
        \textbf{(1) Foundation (early phase):} $\mathcal{L}_{\text{MSE}}$ dominates the total gradient, driving the generator to establish coarse fidelity. 
        \textbf{(2) Transition (middle phase):} $\mathcal{L}_{\text{ADV}}$ progressively increases as the generator shifts from pure reconstruction toward adversarial refinement, improving the statistical plausibility of imputed values. 
        \textbf{(3) Refinement (late phase):} $\mathcal{L}_{\text{ADV}}$ becomes the primary gradient driver, pushing the generated distribution toward statistical indistinguishability from the observed data.
    Throughout all phases, $\mathcal{L}_{\text{OT}}$ maintains a consistent contribution, providing persistent geometric regularization. The convergence of all components to stable ratios indicates that the alternating optimization reaches a stable equilibrium.
    
    \subsection{Sensitivity Analysis} \label{sec:exp_para}
    This subsection examines the sensitivity of CAGI to key hyperparameters (i.e., Loss Weights, Number of Clusters, and Update Frequencies).

    \textbf{Loss Weights.}  
    Figure~\ref{fig:training_ab} presents a grid search over $\alpha \in [0.5, 500]$ and $\beta \in [0.1, 5]$.
    RMSE on \textit{spam} remains near‑optimal across a broad region where $\alpha \geq 10$, with degradation occurring primarily when $\alpha$ is small and $\beta$ simultaneously large; a similar pattern emerges on \textit{letter}. 
    Overall, $\alpha$ must maintain sufficient dominance to anchor instance-level fidelity, while $\beta$ serves as a complementary regularizer whose exact value is not critical within a reasonable range.

\begin{figure*}[t] 
        \centering 
        \includegraphics[width=\textwidth]{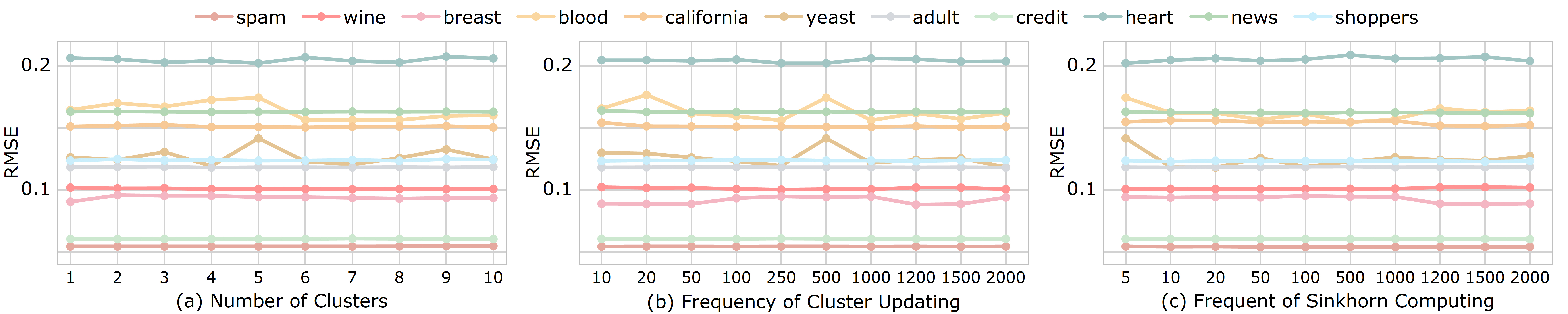}
        \caption{Sensitivity to: (a) number of clusters $K$, (b) cluster update frequency, and (c) Sinkhorn computation frequency. Default setting: $K=5$, cluster update frequency$=500$, Sinkhorn frequency$=5$.} \label{fig:training_para}
    \end{figure*}

    \textbf{Number of Clusters.} 
    As shown in Figure~\ref{fig:training_para}(a), performance is remarkably stable across a wide range of $K$ values. 
    This empirically confirms the design choice discussed in Section~\ref{sec:method_design}: since cluster assignments serve as soft locality priors rather than strict subgroup identifiers, the model tolerates substantial variation in $K$ without significant degradation.

    \textbf{Update Frequencies.}
    The cluster update frequency (Figure~\ref{fig:training_para}(b)) reveals a clear trade-off: very frequent updates 
    introduce instability as the generator continuously adapts to rapidly shifting assignments, while very infrequent updates condition the generator on stale subgroup information. The optimal range of 50--500 iterations provides sufficient time for the generator to adapt to current assignments before receiving updated guidance. 
    The Sinkhorn computation frequency (Figure~\ref{fig:training_para}(c)) exhibits a similar trade-off.

    \begin{figure}[t]
        \centering
        \includegraphics[width=\textwidth]{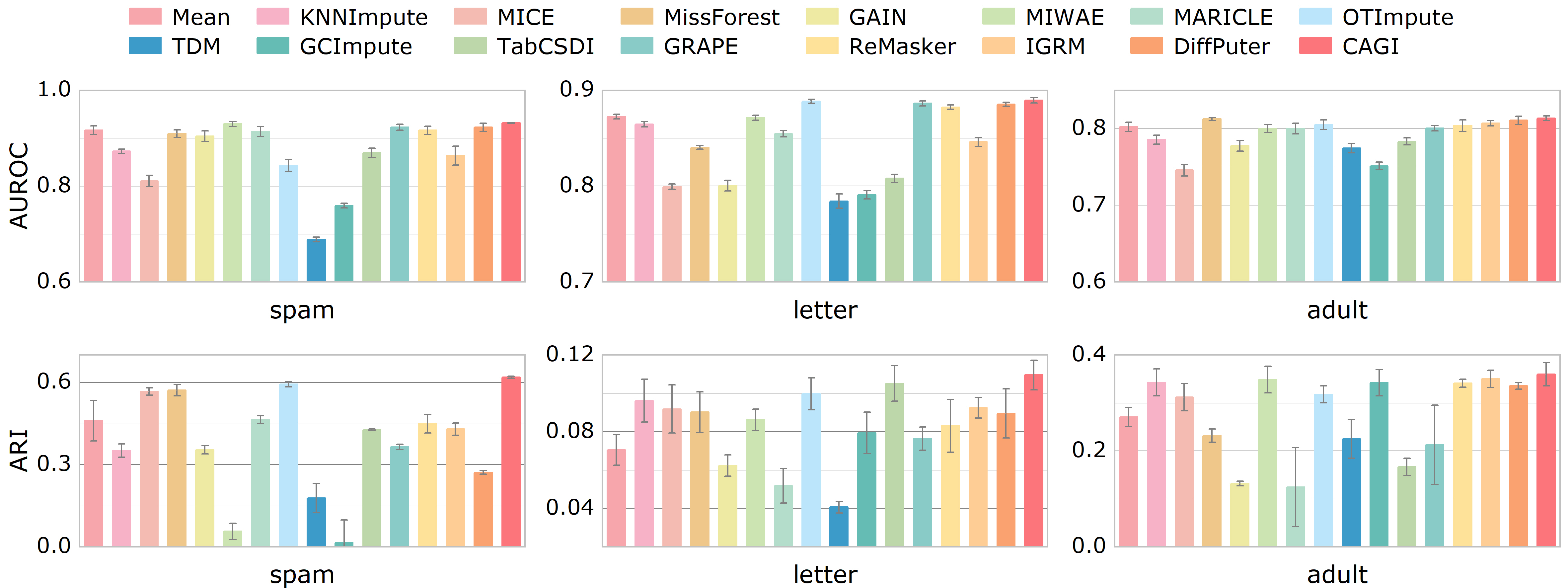}
        \caption{Classification performance (top, measured by AUROC) and clustering performance (bottom, measured by ARI) on \textit{spam}, \textit{letter}, and \textit{adult}. Higher values indicate better performance for both metrics.}
        \label{fig:all}
    \end{figure}
    
    \subsection{Downstream Task Evaluation} \label{sec:dt}
    This subsection evaluates whether CAGI's improved imputation fidelity translates into tangible gains for classification and clustering.
    
    \textbf{Classification Performance.}    
    As shown in Figure~\ref{fig:all}(top), CAGI achieves competitive or superior AUROC scores across all datasets. This indicates that the improved imputation fidelity translates into downstream gains: higher-quality imputations preserve discriminative signals that benefit the classifier.

    \textbf{Clustering Performance.}     
    As shown in Figure~\ref{fig:all}(bottom), CAGI achieves the highest or near-highest ARI on the majority of datasets. This is directly explained by CAGI's design: explicitly modeling subgroup structure during imputation naturally preserves inter-cluster separability. Such preservation is valuable for unsupervised downstream tasks where distributional fidelity matters more than pointwise accuracy.

\section{Conclusion}
    In this paper, we propose CAGI to guide imputation via a ``Partition-Guide-Restore'' strategy. Cluster assignments and imputed values are jointly refined through an iterative feedback loop, where each progressively improves the others. To ensure both instance-level fidelity and distribution-level consistency, CAGI adopts a multi-level learning objective combining cluster-conditioned adversarial generation with optimal-transport regularization.
    Extensive experiments demonstrate CAGI's superiority in imputation fidelity and downstream tasks utility. 
    While CAGI shows robust performance across a range of cluster counts $K$, automatically determining an appropriate $K$ could further tailor the model to the intrinsic structures of different datasets and improve accuracy. Future work will explore data-driven cluster granularity selection, investigate the interplay between subgroup structure and non-random missingness mechanisms, and further generalize the framework to temporal data with sequential dependencies.

\bibliographystyle{splncs04}
\bibliography{mybibliography}

\end{document}